\documentclass[runningheads]{llncs}

\usepackage{eccv}

\usepackage{eccvabbrv}
\usepackage{mathtools}

\usepackage{graphicx}
\usepackage{array}
\usepackage{booktabs}
\usepackage[thinlines]{easytable}
\usepackage{multirow}
\usepackage[accsupp]{axessibility}  % Improves PDF readability for those with disabilities.

\usepackage[pagebackref,breaklinks,colorlinks,citecolor=eccvblue]{hyperref}
\usepackage{orcidlink}
\usepackage{bbm}
\newcommand{\shadow}[1]{}
\newcommand{\blue}[1]{\textcolor{blue}{#1}}
\newcommand{\red}[1]{\textcolor{red}{#1}}

\def\b{\blue}
\def\s{\shadow}
\def\r{\red}

\begin{document}

\title{Forward-Backward Knowledge Distillation for Continual Clustering} 

\titlerunning{Abbreviated paper title}

% TODO FINAL: Replace with your author list. j
\author{Mohammadreza Sadeghi \inst{1,2} \and
Zihan Wang\inst{1,2}\and
Narges Armanfard\inst{1,2}}


\institute{McGill University \and
Mila - Quebec Artificial Intelligence Institute\\
\email{\{Mohammadreza.sadeghi, Zihan.wang5, and Narges Armanfard\}@mcgill.ca}}

\maketitle

\vspace{-5mm}
\begin{abstract}

Unsupervised Continual Learning (UCL) is a burgeoning field in machine learning, focusing on enabling neural networks to sequentially learn tasks without explicit label information. Catastrophic Forgetting (CF), where models forget previously learned tasks upon learning new ones, poses a significant challenge in continual learning, especially in UCL, where labeled information of data is not accessible. CF mitigation strategies, such as knowledge distillation and replay buffers, often face memory inefficiency and privacy issues. Although current research in UCL has endeavored to refine data representations and address CF in streaming data contexts, there is a noticeable lack of algorithms specifically designed for unsupervised clustering. To fill this gap, in this paper, we introduce the concept of Unsupervised Continual Clustering (UCC). We propose Forward-Backward Knowledge Distillation for unsupervised Continual Clustering (FBCC) to counteract CF within the context of UCC.
FBCC employs a single continual learner (the ``teacher'') with a cluster projector, along with multiple student models, to address the CF issue. The proposed method consists of two phases: Forward Knowledge Distillation, where the teacher learns new clusters while retaining knowledge from previous tasks with guidance from specialized student models, and Backward Knowledge Distillation, where a student model mimics the teacher's behavior to retain task-specific knowledge, aiding the teacher in subsequent tasks. FBCC marks a pioneering approach to UCC, demonstrating enhanced performance and memory efficiency in clustering across various tasks, outperforming the application of clustering algorithms to the latent space of state-of-the-art UCL algorithms.

\s{FBCC marks the first approach to UCC, effectively addressing the challenge of CF while maintaining memory efficiency. Experimental results illustrate the efficacy of FBCC in clustering samples from various tasks, outperforming the application of clustering algorithms to the latent space of UCL algorithms.}

\s{Unsupervised Continual Learning (UCL) is a burgeoning field in machine learning, focusing on enabling neural networks to sequentially learn tasks without explicit label information. While existing UCL studies focus on refining representations for data streams, there is a lack of continual learning algorithms explicitly tailored for clustering task. To fill this gap, this paper introduces the concept of Unsupervised Continual Clustering (UCC). Catastrophic Forgetting (CF), where models forget previously learned tasks upon learning new ones, poses a significant challenge in continual learning, and the CF gets even more challenging when dealing with unlabeled data. While various strategies, such as knowledge distillation from models trained on previous tasks and replay buffers, have been proposed to mitigate CF, they often suffer from memory inefficiency or privacy concerns. In this paper, we propose Forward-Backward Knowledge Distillation for unsupervised Continual Clustering tasks (FBCC).
FBCC employs a single continual learner (the ``teacher'') with a cluster projector, along with multiple student models, to address CF. The proposed method consists of two phases: Forward Knowledge Distillation, where the teacher learns new clusters while retaining knowledge from previous tasks with guidance from specialized student models, and Backward Knowledge Distillation, where a student model mimics the teacher's behavior to retain task-specific knowledge for future training iterations. FBCC marks the first approach to UCC, effectively addressing the challenge of CF while maintaining memory efficiency. Experimental results illustrate the efficacy of FBCC in clustering samples from various tasks, outperforming the application of clustering algorithms to the latent space of UCL algorithms.}

\end{abstract}

\vspace{-9mm}
\section{Introduction}
\label{sec:intro}

\begin{figure*}[t]
  \centering
  \includegraphics[width=0.80\linewidth]{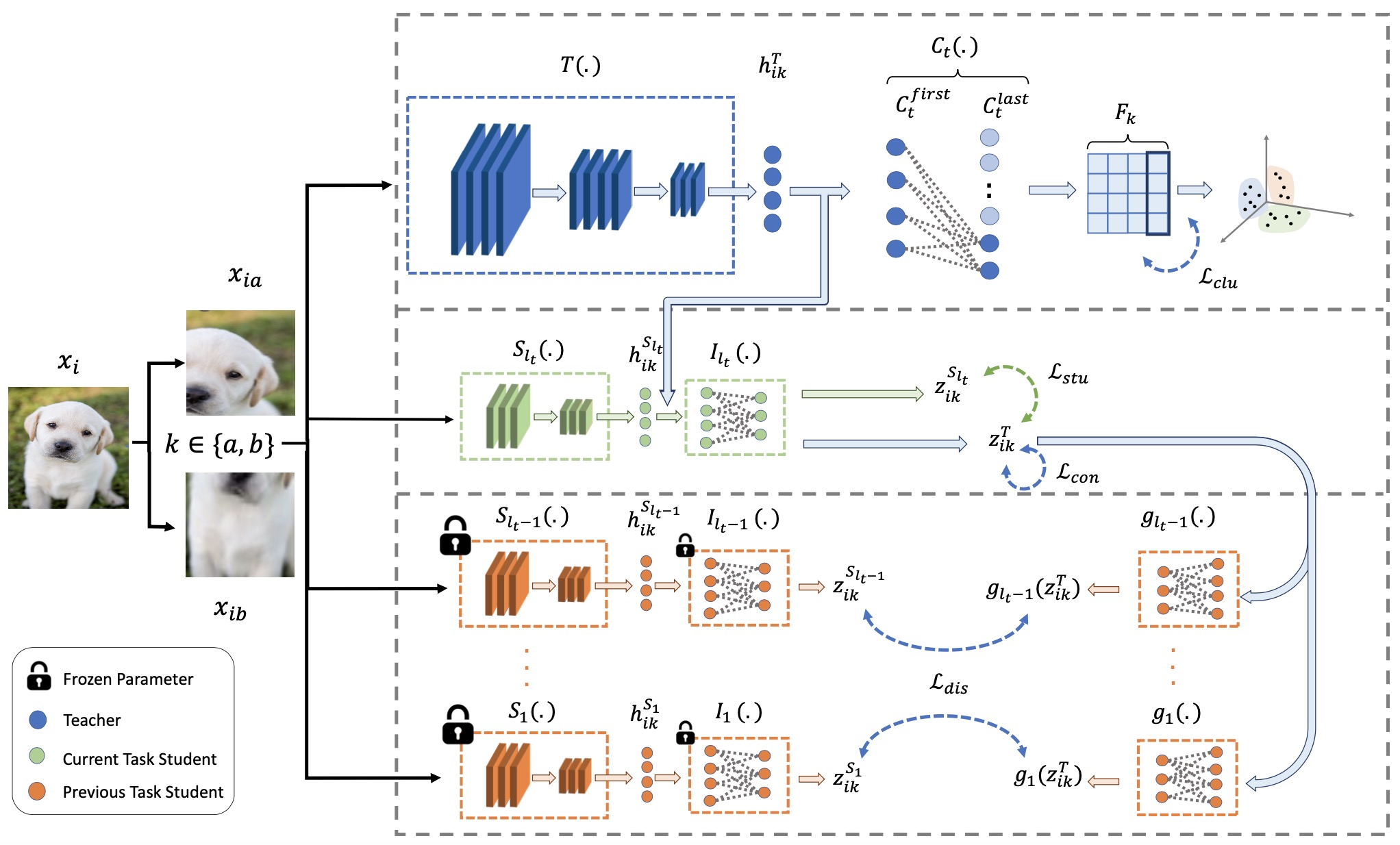}
\caption{Overview of the FBCC Framework for Task $t$: The teacher network, shown in blue, focuses on jointly clustering samples and learning representations for the current task while retaining knowledge of previous tasks with assistance from student networks trained on earlier tasks, depicted in orange. The student network for the current task, shown in green, aims to emulate the teacher network on the current task to assist the teacher in future tasks.}
\label{model}
\end{figure*}

Continual Learning (CL) \cite{wang2023comprehensive} in machine learning endeavors to empower neural networks, to sequentially acquire knowledge across \textit{tasks}. Each task presents a portion of a dataset, inaccessible in its entirety due to constraints such as limited memory, privacy concerns, and concept drift. Within the field of CL, three main approaches emerge: 1- Supervised Continual Learning (SCL) \cite{buzzega2020dark, mai2021supervised}: This approach involves providing the learner with class \textit{labels} for each sample within every task during training iterations. 2- Semi-supervised Continual Learning (SeCL) \cite{wang2021ordisco, lechat2021semi}: This approach aims to leverage the benefits of both supervised learning (using labeled data) and unsupervised learning (using unlabeled data) to improve performance over time while adapting to new tasks or data distributions.  SCL and SeCL predominantly serve the purpose of classifying data samples within each task. 3- Unsupervised Continual Learning (UCL) \cite{cassle,lump}: Unlike SCL and SeCL, UCL focuses on learning new representations for unlabeled data streams evolving over time. These learned representations find utility in subsequent tasks such as data annotation. UCL holds practical significance in real-world applications, given the frequent unavailability of labeled data. While numerous UCL studies, such as those by \cite{lump,cassle,SCALE}, strive to continually refine representations for data, to our knowledge, there exists no continual learning algorithm explicitly tailored for clustering task. In this paper, we introduce the novel concept of Unsupervised Continual Clustering (UCC), specifically designed for clustering tasks. UCC aims to discern clusters within sequentially arriving data across various tasks, with no overlap in cluster contents. This scenario parallels class-incremental continual learning, albeit without access to data sample labels. UCC is ideal for large dataset scenarios, enabling clustering models to identify and adapt to new clusters without forgetting previous ones. For example, it is useful in image annotation and retrieval \cite{chen2020survey,zhou2023remote}, where a clustering model trained on a large dataset encounters a new dataset containing distinct objects or clusters. Offline clustering methods, which retrain models with new and existing data, are memory-intensive and raise privacy concerns. UCC offers an efficient solution by adapting to new data without full retraining or compromising the original dataset's privacy.

\vspace{-0.5mm}
The main challenge of all of the CL approaches is ``Catastrophic Forgetting'' (CF), where learning new tasks leads to the model forgetting previously learned ones \cite{wang2023comprehensive}. In SCL, SeCL, and UCL various strategies have been developed to address CF. Some methods such as \cite{wang2021ordisco} involve training generative model replays and regularizing discriminator consistency to alleviate CF. However, it is worth noting that training a generative model is a resource-intensive and time-consuming process. Another strategy involves storing samples from past tasks in a replay buffer \cite{lump, li2022learning, kiera,buzzega2020dark, mai2021supervised}. However, privacy concerns may preclude the storage of past samples in the buffer. Algorithms such as \cite{cassle, li2022learning} employ a technique known as knowledge distillation (KD) to transfer \textit{insights} from previous tasks. This involves storing models from prior tasks in memory, leveraging them to assist the current model in retaining past knowledge. Nonetheless, this method suffers from memory inefficiency, particularly with deep models containing millions of parameters. Additionally, solutions like \cite{cassle} only retain a single previous task model in memory, potentially leading to forgetting when confronted with numerous tasks.

\s{a neural network aims at 
\r{We need to place our work among other approaches as we are introducing a novel concept to the field of CL. for example: 
1. Define Continual Learning (CL),  Task, Task Label and Class Label and that in this paper we conisider class  incremental learning set up so define it. 4. Define Supervised Continual Learning (S-CL), which has includes a notion of data classification error in its loss, 5. Define Unsupervised Continual Learning (U-CL) that has been only used for representation learning (not specified for classification nor clustering) so far; this category covers self-supervised methods such as Cassle. 6. Define Unsupervised Continual Clustering (UCC), which is unsupervised continual learning tailored for data clustering where no task or class labels are employed.  
- then define CF which is the main challenge in any of the above CL approaches. Then introduce the concept of KD. 
-- then In this paper we propose the first UCC .... All paragraphs till the paragraph starting with ``In This Paper...'' must be revised and re-arranged}}

\s{Clustering algorithms have garnered significant attention within the machine learning literature \cite{oyewole2023data, DML, IDECF, DCSS}. In recent years, contrastive learning-based clustering methods \cite{CC, C3} have shown promising performance in offline training scenarios with large datasets. However, real-world applications often pose a different challenge, where unlabeled data streams progressively over time. In such dynamic environments, continually retraining the offline clustering model on the entire dataset to incorporate new information becomes impractical, expensive, and sometimes impossible if the historical data is inaccessible. This challenge is exacerbated by the computational expense associated with contrastive-learning-based approaches.
\b{In this paper, we introduce the novel concept of Unsupervised Continual Clustering (UCC)}\s{Unsupervised Continual Learning (UCL) \cite{wang2023comprehensive} is a \r{burgeoning} field within machine learning} that explores the capacity of neural networks to learn sequential tasks without \r{explicit WHAT do you mean by explicit} label information. 

\b{In general, the primary challenge in the Continual Learning field} is ``Catastrophic Forgetting'' (CF), where learning new tasks leads to the model forgetting previously learned ones \cite{wang2023comprehensive}. 
\b{In the field of supervised Continual Learning, primarily used for data classification when labeled data are available per learning task,} various strategies have been developed to address CF. One such strategy involves storing samples from past tasks in a replay buffer \cite{lump, li2022learning, kiera}. However, privacy concerns may preclude the storage of past samples in the buffer. Algorithms such as \cite{cassle, li2022learning} employ a technique known as knowledge distillation (KD) to \b{only transfer \textit{insights}} from previous tasks. This involves storing models from prior tasks in memory, leveraging them to assist the current model in retaining past knowledge. Nonetheless, this method suffers from memory inefficiency, particularly with deep models containing millions of parameters. Additionally, solutions like \cite{cassle} only retain a single previous task model in memory, potentially leading to forgetting when confronted with numerous tasks.

\b{Unsupervised} continual clustering pertains to the challenge of identifying clusters within data that arrive sequentially across various tasks. Each task encompasses samples associated with specific clusters, and \b{similar to the supervised CL, } tasks do not overlap in terms of cluster content. To the best of our knowledge, there currently exists no UCL algorithm specifically tailored for clustering tasks\s{, except \cite{he2021unsupervised}. This algorithm initially assigns labels to data samples using a clustering algorithm such as k-means, followed by training a supervised continual learner based on these labels. However, a limitation of this approach is the potential unreliability of labels obtained from the clustering algorithm, which can adversely impact the model's final performance}.}

In this paper, we propose an innovative solution termed \textbf{F}orward-\textbf{B}ackward Knowledge Distillation for mitigating CF in the domain of unsupervised \textbf{C}ontinual \textbf{C}lustering (\textbf{FBCC}). In FBCC, we introduce a single continual learner, also referred to as the ``teacher'', which comprises a deep neural network characterized by a substantial number of parameters. we incorporate a cluster projector designed to map the output of the network to a suitable space tailored for the clustering. Additionally, we develop $M$ student models, with $M$ being a hyperparameter set to a value less than the total number of tasks. Each student model is characterized by significantly fewer parameters in comparison to the teacher model. We propose to train each student to specialize in generating the representations learned by the teacher for a particular task. FBCC has two main phases:
\textbf{1- Forward Knowledge Distillation:} During this phase, we guide the teacher network to simultaneously learn a suitable latent representation for samples within the current task and perform the clustering process. The primary objective is to ensure that the teacher retains the clusters learned from previous tasks while efficiently acquiring knowledge about new clusters in the current task. To achieve this, we propose employing KD from $M-1$ students, each of which is previously specialized to emulate the behavior of the teacher in a particular previous task. 
 \textbf{2- Backward Knowledge Distillation: } In this phase, our proposal involves training a student model with significantly fewer parameters compared to the teacher network. This student model is specifically designed to imitate the behavior of the teacher network, regenerating representations obtained from the teacher for data samples of the current task. Our objective is to store this specialized student model in memory, enabling it to perform knowledge distillation to remind the teacher about the current task during the training of the teacher in future tasks. 
 
 To sum up, the main novelties of this paper are as follows:
 %\vspace{-4.5mm}
\begin{itemize}
    \item  FBCC stands out as the pioneering UCC framework that seamlessly integrates representation learning and clustering, offering a novel solution to the challenge of continual clustering in dynamic environments.
    \item FBCC mitigates CF by utilizing knowledge distillation from specialized student models to guide the teacher network, ensuring retention of past knowledge while learning new tasks.
    \item FBCC introduces a novel knowledge distillation phase where a student model, with significantly fewer parameters, mimics the behavior of the teacher network. This process aids in retaining task-specific knowledge for future tasks.
    \item FBCC offers a memory-efficient approach to continual clustering, where task-specific knowledge is stored in specialized light-weight student models rather than storing samples or large-scale models from past tasks, thereby overcoming memory inefficiency and privacy concerns associated with existing methods.
\end{itemize}
\s{learn the cluster assignments for the samples in the current task and  teach the teacher to obtain a suitable representation for samples from the current task that are far from samples from clusters learned in previous tasks.  

and also retain the knowledge from previous tasks. 

We aim at teaching teacher to learn clusters of samples from the new task and remember clusters obtained from previous tasks by getting help from students. In this paper, we propose two novel KD approaches one from teacher to student to teach the student the current clusters and another from students that are trained on the previous tasks to the teacher to mitigate catastrophic forgetting. At the end of training,  

the continual learner is a teacher that teaches the current task to a student and obtain

You might perform UCL and then apply k-means \cite{kmeans} or spectral clustering \cite{ng2001spectral} to define clusters.}

\vspace{-5mm}
\section{Related Work}
\vspace{-2mm}
\textbf{Unsupervised Continual learning}: The primary challenge in CL lies in combating CF, a phenomenon characterized by neural networks rapidly losing performance on previously learned tasks when presented with new tasks. Traditional continual learning strategies typically address CF through three main approaches: adding explicit regularization \cite{ Jung2020nips, Paik2019OvercomingCF, Aljundi_2018_ECCV}, parameter isolation, and dynamic architecture \cite{rusu2022progressive, Lee2020A, pmlr-v232-doan23a,  Rypes2024iclr, Wang_2023},  and experience replay via a memory buffer \cite{Cha_2021_ICCV, De_Lange_2021_ICCV, arani2022learning,  img_ret, yoon2022online}. However, these methodologies inherently presuppose the availability of task-specific and class-specific labels within the incoming data streams. Conversely, unsupervised continual learning, without task and class labels, presents a significantly more complex challenge. Several methods have been developed to address catastrophic forgetting, with variational autoencoders and generative replay being common approaches \cite{Achille2018LifeLongDR, RAMAPURAM2020381, Rao2019ContinualUR, Wu2018MemoryRG}. Innovations like STAM \cite{stam} employed an expandable memory architecture designed for processing single-pass data streams by incorporating  novelty detection and memory update. LUMP \cite{lump} enhances memory retention by augmenting data and blending new samples with existing ones. CaSSLe \cite{cassle} transforms the self-supervised loss into a knowledge distillation approach by associating the present state of a representation with its preceding state. SCALE \cite{SCALE} addresses CF for non-iid and single-pass data using a self-supervised forgetting loss and online memory update mechanisms. POCON \cite{Gomez_Villa_2024_WACV} trains an expert network solely for new task, followed by an adaptation-retrospection phase to prevent forgetting. Evolve \cite{Yu_2024_WACV} leverages multiple pretrained models as cloud-based experts to enhance existing self-supervised learning methods on local clients.
\\
\\
\textbf{Deep Clustering:} Traditional deep clustering techniques \cite{wang2017research, 6976982} typically involve finding a suitable latent representation (e.g., autoencoder (AE)) for data samples and subsequently employing a clustering algorithm, such as k-means, to assign the latent representation of samples to different clusters. Incorporating representation learning and clustering losses has been shown to yield better performance than traditional deep clustering methods in several studies \cite{IDEC, IDECF, DCN, DSL}. For example, DEC \cite{DEC} utilizes an autoencoder (AE) to jointly assign samples to clusters and refine representations based on the cluster assignments. \cite{IDEC,IDECF} enhance the performance of DEC by incorporating the reconstruction loss of the autoencoder into the loss function introduced by DEC. \cite{DML} employs a general autoencoder for instances that are easily clustered along with separate AEs for difficult-to-cluster data to improve performance. \cite{JULE} performs agglomerative clustering to learn data representation and cluster assignments. In \cite{IIC}, the mutual information between the cluster assignments of a pair is maximized. In recent years, contrastive learning has emerged as a pivotal technique within the realm of unsupervised representation learning \cite{albelwi2022survey,SimCLR,zbontar2021barlow}. Many deep clustering methodologies have seamlessly integrated contrastive learning, resulting in substantial improvements in deep clustering performance. CC \cite{CC} executes contrastive learning at both the instance and cluster levels. \cite{C3} enhances the performance of CC by incorporating inter-sample relationships within the latent space of CC.  \cite{dang2021doubly} devise contrastive loss functions from both sample and class perspectives, thereby fostering the learning of more discriminative representations.

\s{\r{Contrastive learning is not that important to start off with it! you should start with U-CL, then contrastive clustering.}
\textbf{Contrastive Learning:} In recent years, contrastive learning has emerged as a pivotal technique within the realm of unsupervised representation learning \cite{albelwi2022survey,SimCLR,zbontar2021barlow}. The fundamental objective of contrastive learning is to transform original samples into a subspace by minimizing the discrepancy between pairs of positive samples while simultaneously maximizing the gap between pairs of negative samples. \cite{he2020momentum} recognizing contrastive learning as akin to dictionary search, integrates queues for storing additional negative samples, thereby enhancing performance. \cite{SimCLR} generates transformed versions of input samples and aims to learn representations where augmentations of the same data point are brought closer together while being pushed further away from other points. \cite{zbontar2021barlow} explores an objective function that evaluates the cross-correlation matrix between features, while \cite{bardes2021vicreg} incorporates a combination of variance, invariance, and covariance regularizations. Many deep clustering methodologies have seamlessly integrated contrastive learning, yielding substantial improvements in deep clustering performance.  Contrastive Clustering (CC) \cite{CC} pioneers the incorporation of labels as representations, executing contrastive clustering at both the instance and cluster levels. 
\cite{C3} enhances the performance of CC by incorporating the inter-sample relationships within the latent space of CC. \cite{dang2021doubly} devises contrastive loss functions from both sample and class perspectives, thereby fostering the learning of more discriminative representations.}

\s{\textbf{Continual learning paradigms}. In the realm of continual learning, the paradigms of Class-Incremental Learning (CIL) and Task-Incremental Learning (TIL) stand out for their relevance to real-world applications \cite{wang2024comprehensive, vandeVen2018GenerativeRW, hsu2019reevaluating}. CIL challenges models to learn new classes over time without forgetting previous ones, crucial for systems that encounter evolving data landscapes. TIL, in contrast, focuses on learning distinct tasks sequentially with clear task boundaries, necessitating strategies that enable task-specific learning without interference. Our work primarily investigates \textbf{TIL} and \textbf{CIL}, exploring strategies to mitigate catastrophic forgetting in diverse scenarios.

\textbf{Knowledge Distillation}
Knowledge distillation has emerged as a pivotal technique in continual learning. This strategy leverages the concept of transferring nuanced insights from a well-trained model, known as the ``teacher'', to a less experienced ``student'' model. The core objective is to replicate the teacher's performance in the student model by aligning their outputs, be it through logits, output distributions, intermediate layer activations, or attention mechanisms \cite{zagoruyko2017paying, hinton2015distilling, NIPS2014_ea8fcd92}. In the conventional CIL, the model previously trained on old tasks assumes the role of the teacher, guiding the student model that integrates new task data from a baseline established by the teacher's pre-trained weights \cite{Zhao2019MaintainingDA, Wu_2019_CVPR}. Techniques like iCaRL utilize sigmoid outputs for past classes, while others apply normalized softmax outputs with temperature adjustments \cite{Rebuffi_2017_CVPR}. Methods like UCIR enhance memory retention by aligning feature embeddings through cosine similarity \cite{Hou_2019_CVPR}. PODNet, by focusing on pooled feature differences rather than direct comparisons, offers a flexible distillation approach conducive to CIL \cite{PODNet}. Kang et al.~\cite{Kang_2022_CVPR} assess the link between updates in model representations and consequent loss escalations.

(Talk about how difference between our Knowledge Distillation with others)}

% However, these techniques often struggle with scalability and high computational demands when applied to large datasets.
\vspace{-3mm}
\section{Proposed Method}
\vspace{-1mm}
We introduce the UCC problem\s{, denoted as $\Omega$,}  which involves learning a sequence of $N$ tasks. Here, the set of tasks is denoted as $\Omega = \{\mathcal{D}_1, \mathcal{D}_2,...,\mathcal{D}_N\}$, where $\mathcal{D}_t$ corresponds to the dataset from task $t$ ($1\leq t \leq N$). In the UCC setting, although labeled information is not available, it is known that data samples belonging to different tasks are from distinct classes.  In other words, if $\mathcal{Y}_i$ and $\mathcal{Y}_j$ respectively represent sets of class labels from task $i$ and task $j$ ($i \neq j$), then $\mathcal{Y}_i \cap \mathcal{Y}_j = \varnothing$. This scenario is similar to unsupervised class-incremental setting \cite{cassle, wang2023comprehensive}. Following the clustering literature where the number of clusters is predetermined, we assume that the number of clusters present in every task is known in the UCC set-up. The number of cluster for task $t$ is denoted by $\lambda_t$. 

\s{The unsupervised continual learning problem, denoted as $\Omega$, is defined as the process of learning a sequence of $N$ tasks $\Omega = \{\mathcal{D}_1, \mathcal{D}_2,...,\mathcal{D}_N\}$, where $\mathcal{D}_t$ corresponds to the dataset from task $t$ ($1\leq t \leq N$). In an unsupervised continual learning setting, although labeled information is not available, it is known that data samples belonging to different tasks are from distinct classes \r{CITE}. In other words, if $\mathcal{Y}_t$ and $\mathcal{Y}_j$ respectively represent sets of class labels from task $t$ and task $j$, then $\mathcal{Y}_t \cap \mathcal{Y}_j = \varnothing$. In this paper, \b{Following the clustering literature where the number of clusters is predetermined, we assume that the number of classes present in every task is known in the continual clustering set-up.}\s{ we assume that we possess knowledge of the number of classes present in each task.} The number of classes for task $t$ is denoted by $\lambda_t$. }

  \s{For each task $t$, where $1\leq t\leq \text{N}$, a dataset corresponding to task $t$ is represented by $\mathcal{D}_t$, which contains images belonging to a specific set of class labels denoted as $\mathcal{Y}_t$. It's important to note that $\mathcal{Y}_t \cap \mathcal{Y}_j = \varnothing$ for all other tasks $j$, where $j \neq t$. Furthermore, it's essential to highlight that labels are only used for dataset creation, and the set of labels is not utilized during the training process. We only possess knowledge of the number of clusters present in each task. The number of clusters for task $t$ is denoted by $\lambda_t$.}

Given the recent success of contrastive clustering methods, we propose training FBCC's networks using losses inspired by contrastive learning.  In line with other contrastive learning methodologies, we employ two sets of augmentations ${a,b}$ on each sample within the dataset $\mathcal{D}_t =\{x_1,...,x_{|\mathcal{D}_t|}\}$ at task $t$, yielding $\mathcal{D}_{ta} =\{x_{1a},...,x_{|\mathcal{D}_t|a}\}$ and  $\mathcal{D}_{tb} =\{x_{1b},...,x_{|\mathcal{D}_t|b}\}$ respectively.

As depicted in Figure \ref{model}, our approach involves training a deep teacher encoder denoted as $T(.)$. The primary function of $T(.)$ is to derive a suitable clustering-friendly latent representation for the current task while also retaining knowledge from the previous tasks. The latent representations of $T(.)$ for $x_{ik}$ is denoted as $h_{ik}^{T} = T(x_{ik})$. Hereafter, we use $k \in \{a,b\}$ to represent different augmentations\s{ throughout this paper}.

In Section \ref{changing_m}, we demonstrate that considering all previous tasks in training the current teacher is not essential. We propose training up to $M$ light-weight student encoders, where $M$ is a hyperparameter with $1 < M \leq N$, to alleviate catastrophic forgetting. Specifically, when $t < M$, we train $t$ students, and when $t \geq M$, we retain the last $M$ students while removing all others from memory. Henceforth, we denote the number of students for task $t$ as $l_t = \mathbbm{1}\{t < M\} t + \mathbbm{1}\{t \geq M\} M$, where $\mathbbm{1}\{.\}$ is an indicator function. The $r$-th student encoder is denoted as $S_r(.)$, where $1\leq r \leq l_t$. The latent representations of the $r$-th student encoder for $x_{ik}$ is denoted as $h_{ik}^{S_r}=S_r(x_{ik})$. Hereafter, we utilize the notation $r$ to denote indices ranging from 1 to $l_t$, i.e. $1\leq r\leq l_t$. 

To reduce dimensionality and define our contrastive loss, we employ $l_t$ instance-level projectors. The $r$-th instance-level projector is denoted by $I_r(.)$. For $1 \leq r \leq l_t-1$, the $r$-th instance-level projector aims to map $h_{ik}^{S_r}$ to obtain $z_{ik}^{S_r} = I_r(h_{ik}^{S_r})$. We propose to share the $l_t$-th instance-level projector between the $l_t$-th student and the teacher \s{network}. Hence, $I_{l_t}(.)$ is responsible for creating $z_{ik}^{S_{l_t}} = I_{l_t}(h_{ik}^{S_{l_t}})$ and $z_{ik}^{T} = I_{l_t}(h_{ik}^{T})$.

In the following sections, we delve into our forward-backward knowledge distillation approach for continual clustering (FBCC). Firstly, we introduce forward distillation, where our teacher is trained to learn the new task $t$ while leveraging the knowledge of previous tasks present in the previously trained student networks, i.e. students from 1 to $l_t-1$; this is to retain the memory of previous tasks. In the forward mode, all parameters of the students are frozen, and our focus lies solely on training the teacher encoder. Conversely, in the backward mode, we propose a novel approach for training the $l_t$-th student, enabling it to grasp the latent representation of the teacher encoder for task $t$. In the backward mode, all parameters of students ranging from 1 to $l_{t-1}$, as well as the teacher encoder, are frozen, and our attention is directed towards updating the parameters of student $l_t$.
\vspace{-3mm}
\subsection{Forward Knowledge Distillation}
\textbf{Training Teacher on the Current Task:} 
Since we lack access to labeled information for dataset $\mathcal{D}_t$, inspired by \cite{CC,SimCLR}, we train our teacher encoder on samples $x_{ia}$ and $x_{ib}$, which are two augmentations of the same sample $x_i \in \mathcal{D}_t$, in order to maximize the similarity between $z_{ia}^T$ and $z_{ib}^T$, while minimizing the similarity of $z_{ia}^T$ and $z_{ib}^T$ with the remaining $2|\mathcal{B}|-2$ samples in the batch $\mathcal{B}$, where $|\mathcal{B}|$ denotes the batch size.

Moreover, to further segregate representations of the current task from those of previous tasks, given the absence of datasets from previous tasks, we propose ensuring that augmentations of the current task $t$ are distinctly distant from prototypes learned from previous tasks ranging from the first task up to task $t-1$. We discuss the way of defining prototypes later in this section. We present the set of prototypes from the first task up to task $t-1$ as $\mathcal{P}_{t-1}$. This approach facilitates the preservation of task-specific information and aids in reducing interference between tasks, thereby enhancing the model's performance on sequential learning tasks. We propose the following contrastive loss function (i.e. $\mathcal{L}_{con}$) for training our teacher encoder at task $t$.
\begin{align}\label{eq_con}
    \mathcal{L}_{con} = \frac{1}{2|\mathcal{B}|} \sum_{x_i\in \mathcal{B}}(\ell_{ia}^{con} + \ell_{ib}^{con})
\end{align}
\vspace{-3mm}
\begin{align}\label{lossa_con}
    \ell_{ia}^{con}= -\log(\frac{\exp(sim(z_{ia}^{T},z_{ib}^{T}))}{\sum_{j \in \mathcal{B}}^{j\neq i}\sum_{k\in\{a,b\}}\exp(sim(z_{ia}^{T},z_{jk}^{T}))+\sum_{z_p \in \mathcal{P}_{t-1}}\exp(sim(z_{ia}^{T},z_p))})
\end{align}
In this study, $sim(.)$ denotes the cosine similarity between two vectors. Analogous to $\ell_{ia}^{con}$, we define $\ell_{ib}^{con}$, which assesses the similarity between $z_{ib}^{T}$ and other samples in the batch as well as the prototypes set.
\\
\\
\textbf{Knowledge Distillation from Students to Teacher:}
In task $t$, since we do not have access to $\mathcal{D}_1,..., \mathcal{D}_{t-1}$, we propose to utilize $l_t-1$ student networks trained on the last $l_t-1$ tasks to aid our network in not forgetting the previous tasks. 
Our objective is to ensure that $z_{ik}^{T}$ contains at least as much information as (and ideally more than) all $z_{ik}^{S_r}$, where $1 \leq r \leq l_t-1$. Instead of enforcing similarity between  $z_{ik}^{T}$ and $z_{ik}^{S_r}$, which could discourage the new model from learning new concepts, we draw inspiration from \cite{cassle} and propose to map $z_{ik}^{T}$ using fully connected predictor networks $g_r(.)$ to the previous task learned by the $r$-th student (e.g. $z_{ik}^{S_r}$), while freezing the parameters of the $r$-th student and $r$-th instance-level projector. Given the well-studied effectiveness of utilizing contrastive loss between the output of the predictor and the latent representation of the previous task for the current dataset in \cite{cassle}, we define our loss \s{in a similar manner} as follows:
\begin{align}
    \mathcal{L}_{dis} = \frac{1}{2|\mathcal{B}|} \sum_{x_i\in \mathcal{B}}(\ell_{ia}^{dis} + \ell_{ib}^{dis}),
\end{align}
\vspace{-3mm}
\begin{align}
    \ell_{ia}^{dis}= - \frac{1}{l_t-1}\sum_{1 \leq r<l_t}\log(\frac{\exp(sim(g_r(z_{ia}^{T}),\Delta(z_{ia}^{S_{r}})))}{\sum_{j \in \mathcal{B}}^{j\neq i}\sum_{k\in\{a,b\}}\exp(sim(g_r(z_{ia}^{T}),\Delta(z_{jk}^{S_r})))}),
\end{align}
where $\Delta(.)$ denotes the detaching operation, in which we detach vectors from a network and do not have a backward path to this network from our loss. With this approach, our teacher network aims to imitate the behavior of students, which serve as estimations of the previous teacher. The primary distinction between our proposed distillation framework and \cite{cassle} lies in our approach to addressing the catastrophic forgetting issue. We train multiple light-weight students, each capable of serving as a reliable estimation of our teacher network, allowing us to remember more than one previous task while storing a lower number of parameters in memory. In contrast, \cite{cassle} relies solely on the large-scale deep network learned in the previous task, potentially leading to forgetting of the initial tasks when confronted with numerous tasks. Moreover, in Section \ref{ablation}, we elucidate the significance of retaining memories of more than one previous task in mitigating the catastrophic forgetting issue. 
\\
\\
\textbf{Clustering Samples of the Current Task: }  Inspired by \cite{CC}, for clustering samples belonging to $\mathcal{D}_t$, we propose to train a cluster-level projector network for task $t$, denoted by $C_t(\cdot)$. This network consists of a 2-layer fully connected network followed by a softmax function, which maps the latent representation of the teacher encoder to a suitable space designed for the clustering task. The first layer and the last layer of $C_t(\cdot)$ are denoted as $C_t^{\text{first}}$ and $C_t^{\text{last}}$, respectively, i.e.  $C_t = C_t^{\text{last}}(C_t^{\text{first}})$. At the beginning of task $t$, where $2\leq t \leq N$, we initialize $C_t^{\text{first}}$ with $C_{t-1}^{\text{first}}$.  This initialization is intended to retain the information from previous tasks during this stage. The output of $C_t^{\text{first}}$ for sample $x_{ik}$ is denoted by $\hat{h}_{ik} = C_t^{\text{first}} (T(x_{ik}))$. Also, we propose to store $C_{t-1}^{\text{last}}$ in memory for use during the test phase. Within $C_t^{\text{last}}$, we allocate $\lambda_t$ neurons to transform $\hat{h}_{ik}$ into a specialized space designed for the clustering of data samples within the current task. We initialize $C_t^{\text{last}}$ with random values. For instance, if we assume 2 clusters per task, in Figure \ref{model}, for the new task, we add two new neurons, shown in dark blue, to create $C_t^{\text{last}}$. Also, we store neurons of previous tasks shown in light blue in memory. 

In every task, for a batch of data $\mathcal{B}$, we create two augmentations of the batch to obtain $\mathcal{B}_k$, where $k \in \{a,b\}$. We then pass these two augmented batches to the teacher network and the cluster-level projector to obtain $F_k=C_{t}(T(\mathcal{B}_k))$, where $F_k = [f_{1k}| f_{2k}|...|f_{\lambda_tk}]\in \mathbbm{R}^{|\mathcal{B}|\times \lambda_t}$, and $f_{jk} \in \mathbbm{R}^{|\mathcal{B}|}$ represents the probability vector for assigning samples from $\mathcal{B}_k$ to cluster $j$. Inspired by \cite{CC}, we apply contrastive loss on the features of $F_k$ (e.g., $f_{jk}$) instead of applying the contrastive loss between samples. The motivation stems from the derivation of $F_a$ and $F_b$ from two augmentations of the same batch. Therefore, similar clusters represented in $F_a$ and $F_b$ (e.g. $f_{ia}$ and $f_{ib}$) are expected to possess matching probability assignments and ideally be situated far apart from dissimilar clusters. This strategy is designed to promote distinct and well-separated clusters, thereby improving the overall quality of the clustering process. The loss is defined as follow:
\begin{align}
    \mathcal{L}^{clu} = \frac{1}{2\lambda_t}\sum_{i=1}^{\lambda_t}(\ell_{ia}^{clu} + \ell_{ib}^{clu}) - H(F),
\end{align}
\vspace{-4mm}
\begin{align}
    \ell_{ia}^{clu}= - \log(\frac{\exp(sim(f_{ia},f_{ib}))}{\sum_{j=1\, j\neq i}^{\lambda_t}\sum_{k\in\{a,b\}}\exp(sim(f_{ia},f_{jk})}), 
\end{align}
where $H(F) = \sum_{k \in \{a,b\}}\sum_{j=1}^{\lambda_t} -\mathcal{Q}(f_{jk})\log(\mathcal{Q}(f_{jk}))$ is entropy of cluster assignments probabilities, where $\mathcal{Q}(f_{jk}) = ||f_{jk}||_1/||F_k||_1$ and $||.||_1$ denotes the $\ell_1$ norm. We maximize the entropy to avoid the trivial solution of converging all assignments to one cluster. 
\\ \\
\textbf{Updating Prototype Set:} At the end of training of task $t$, we propose to define $\lambda_t$ new prototypes that are representatives of task $t$. These prototypes, denoted as $p_v$ for $1 \leq v \leq \lambda_t$, are intended for use in \eqref{eq_con} where we generate representations of future tasks to be distinct from the representations learned during the current task. $p_v$ is designed to maintain the same distance with samples belonging to the $v$-th cluster.\s{Furthermore, to enhance the quality of prototypes for the clusters of the current task, we only consider samples if both of its augmentations are assigned to }\s{we limit consideration to only two sample augmentations when they originate from the same cluster ... NOT CLEAR WRITING.} Given $c_{ia}$ and $c_{ib}$ as the cluster assignments of $x_{ia}$ and $x_{ib}$ (i.e. $c_{ia} = \text{argmax}[C_t(T(x_{ia})]$ and $c_{ib} = \text{argmax}[C_t(T(x_{ib})]$), we can formulate the following equation for determining $p_v$.

\begin{align}
    p_v = \frac{\sum_{x_i \in \mathcal{B} } \mathbbm{1}\{c_{ia} = v\, \, \text{and}\, \, c_{ib} = v\}(z_{ia}^T + z_{ib}^T)}{2\sum_{x_i \in \mathcal{B} } \mathbbm{1}\{c_{ia} = v\, \, \text{and}\, \, c_{ib} = v\}}
\end{align}
To enhance the quality of prototypes for the clusters of task t, the prototype of the $v$-th cluster is the center of the ``reliable'' augmented samples in the z space. We consider a sample as ``reliable'' if both of its augmentations are assigned to the same cluster. Once new prototypes have been identified for task $t$, we incorporate them into the existing set of prototypes obtained from preceding tasks $\mathcal{P}_{t-1}$ to constitute the updated prototype set $\mathcal{P}_t$.
\vspace{-4mm}
\subsection{Backward Knowledge Distillation}
In task $t$, we propose training a light-weight student encoder with significantly fewer parameters compared to our teacher encoder, with the aim of replicating the behavior of the teacher encoder on task $t$. Our objective is to store this student model in memory for use in knowledge distillation for future tasks. For each task, we limit the memory storage to at most $M$ student encoder networks, ensuring that the total number of parameters across these networks is less than that of our teacher network. This approach enhances memory efficiency compared to methods such as \cite{cassle}, which necessitate storing their large-scale network from the current task in memory for knowledge distillation in subsequent tasks. In task $t$, we propose to train the $l_t$-th student network with the following loss function, while keeping the parameters of the teacher encoder and all other student encoders frozen.
\vspace{-3mm}
\begin{align}
    \mathcal{L}_{stu} = \frac{1}{2|\mathcal{B}|}\sum_{x_i\in \mathcal{B}} (\ell_{ia}^{stu} + \ell_{ib}^{stu})
\end{align}
\vspace{-4mm}
\begin{align}\label{loss_stu}
    \ell_{ia}^{stu}= \frac{1}{|h_{ia}^{S_{l_t}}|}|| h_{ia}^{S_{l_t}}-\Delta(h_{ia}^{T})||_2^2-\log(\frac{\exp(sim(z_{ia}^{S_{l_t}},\Delta(z_{ia}^{T})))}{\sum_{j \in \mathcal{B}}^{j\neq i}\sum_{k\in\{a,b\}}\exp(sim(z_{ia}^{S_{l_t}},\Delta(z_{jk}^{T})))})
\end{align}

Where $||.||_2$ represents the $\ell_2$ norm and $|h_{ia}^{S_{l_t}}|$ shows the number of elements in $h_{ia}^{S_{l_t}}$. With this loss function, we aim to instruct our student network in two critical aspects: 1- Our student must grasp the representations produced by the teacher network irrespective of other samples. This is achieved through the first component of our loss. 2- Our student network must follow the same structural relationships established by the teacher encoder within a batch; we propose to enforce such behavior through defining a contrastive loss between the student and teacher network outputs, as is shown in the second term of the loss defined in \eqref{loss_stu}.\s{enforce the similarity between the student and teacher network outputs using a contrastive loss.} These two components of the loss function enable the $l_t$-th student to learn both the output and the relationships between samples produced by the teacher encoder.
\vspace{-4mm}
\subsection{Overall Training Scheme}
For the batch $\mathcal{B}$ comprising data samples from task $t$, we initially fix the parameters of all students and conduct forward distillation to minimize the combined losses of contrastive, distillation, and clustering (i.e., $\mathcal{L}_{con} + \mathcal{L}_{dis} + \mathcal{L}_{clu}$). Subsequently, we proceed with backward distillation, wherein we freeze the parameters of the teacher and unfreeze the parameters of the $l_t$-th student, minimizing $\mathcal{L}_{stu}$ for the same batch $\mathcal{B}$. Figure \ref{model} shows the overall training scheme for task $t$.  Furthermore, Algorithm 1 in the Supplementary Material file provides the pseudo-code for the training of FBCC.
\\ \\
\textbf{Assigning Samples to Clusters:} For the UCC setting, after completing training on the last task $N$, to find the final cluster assignments for sample $x_i$, we propose to use the trained teacher encoder (i.e., $T$), the first layer of the cluster projector (i.e., $C_N^{\text{first}}$), which is shared among all tasks, and the last layers of the cluster projector, which are task-specific and stored in memory (i.e., $C_1^{\text{last}},..., C_N^{\text{last}}$). To assign cluster label $c_i$ to data sample $x_i$, we propose to obtain the latent representation of data $\hat{h}_i = C_N^{\text{first}}(T(x_i))$, then map $\hat{h}_i$ to different clustering spaces using the last layers trained for each task and pick the index of the maximum value, i.e., $c_i = \text{argmax}[C_1^{\text{last}}(\hat{h}_i),..., C_N^{\text{last}}(\hat{h}_i)]$.

\s{\textbf{Task-Incremental:} In the unsupervised task-incremental setting, we have an additional information about task ID of each sample in the test phase. We show the task ID for sample $x_i$ by $\mathcal{Y}^{task}_i$. In task-incremental setting, instead of using teacher to obtain cluster label for sample $x_i$, we proposed to use $M$ students for assigning samples to different clusters. $M$ students are trained to imitate the behavior of the teacher for the last $M$ tasks. Also, since the teacher learn to }
\vspace{-3mm}
\section{Experiments and Results}

In this section, we conduct comprehensive experiments to illustrate the effectiveness of our proposed method. We assess our model's performance on three challenging computer vision benchmark datasets: CIFAR-10 \cite{Cifar} (with 10 classes and 5 tasks), CIFAR-100 \cite{Cifar} (with 100 classes and 10 tasks), and Tiny-ImageNet \cite{Imagenet} (with 200 classes and 10 tasks). \s{the datasets?? IT IS NOT!}To train our model, we concatenate the train and test sets of the datasets, a common practice in clustering research (e.g., \cite{DCSS, DML, CC, C3}).

We evaluate the performance of our clustering model using two key metrics: average clustering accuracy ($\overline{\text{ACC}}$) and average forgetting ($\overline{\text{F}}$), where ACC \cite{acc} is a widely used metric for assessing clustering performance and average forgetting is a common metric used to measure how much information the model has forgotten about previous tasks. $\overline{\text{ACC}}$ and $\overline{\text{F}}$ are defined as follows:
\begin{align}
    &\overline{\text{ACC}} = \frac{1}{\text{N}} \sum_{t=1}^{\text{N}} \text{ACC}_{t,\text{N}}\\
    \overline{\text{F}} &= \frac{1}{\text{N}-1} \sum_{i=1}^{\text{N}-1} \max_{t \in \{1,.., \text{N}-1\}} (\text{ACC}_{i,t}-\text{ACC}_{i,\text{N}}),
\end{align}
where $\text{ACC}_{i,j}$ is $\text{ACC}$ of task $i$ at the end of training of task $j$.  Implementation details and additional experiments are presented in the Supplementary Material file.
\s{Additionally, we present an ablation study on our proposed method in \r{Section}. Furthermore, in \r{Section}, we demonstrate how our proposed framework can be adapted to a task-incremental setting and provide experimental results showcasing the performance of our model.}
\\
\\
\begin{table}[htbp]
  \caption{FBCC Performance Comparison in terms of $\overline{\text{ACC}} (\%)$ and $\overline{\text{F}} (\%)$. The best result for continual learning algorithms in each column is highlighted in bold.}
    \centering
    \setlength{\tabcolsep}{5pt}
    \renewcommand{\arraystretch}{1.5}
    \scalebox{0.9}{\begin{tabular}{|c|c|c|c|c|c|c|c|}
        \hline
        \multirow{2}{*}{Algorithms} &
         \multicolumn{2}{c|}{CIFAR-10} & \multicolumn{2}{c|}{CIFAR-100} & \multicolumn{2}{c|}{Tiny-ImageNet}\\
        \cline{2-7}
       & $\overline{\text{ACC}}$ $(\uparrow)$ & $\overline{\text{F}}$ $(\downarrow)$ &$\overline{\text{ACC}}$ $(\uparrow)$ & $\overline{\text{F}}$ $(\downarrow)$ &$\overline{\text{ACC}}$ $(\uparrow)$ & $\overline{\text{F}}$ $(\downarrow)$\\
       \hline
       CC (offline) & 79.00 & - & 42.90 & - & 14.00 & -\\
        \hline \hline
        Co$^2$L (SCL) & 28.35 & 14.05 & 19.88 &  10.30 & 8.69 & 4.95\\
        OCD-Net (SCL) & 40.41 & 7.03 & 18.06 &  7.46 & 7.97 & 5.31\\
        CCL & 36.56 & 6.21 & 19.59& 8.51 & 8.21 & 4.68\\
        STAM & 39.61  &  5.15 & 25.34  & 6.25 & 9.21 & 4.26 \\
        LUMP & 56.43 & 12.76 & 19.53& 6.16 & 10.53 & 2.51\\
        CaSSLe & 40.56 & 3.28 & 36.67 & 3.92 & 17.45 & 2.69\\
        
        \hline \hline
        % FBCC w/o Pro & 75.00 & 2.19 & 37.61 & 4.10 & 17.91 & 2.37\\
        % FBCC w/o KD & 67.54 & 9.21 &  32.47  & 13.31 & 14.28 &6.58\\
        % FBCC + CaSSLe & 70.69  & 4.63 & 35.21  & 6.41& 15.28& 3.01\\
        \textbf{FBCC} & \textbf{75.35} & \textbf{2.12} &   \textbf{38.33} & \textbf{3.81}& \textbf{18.25} & \textbf{2.03}\\
       \hline 
        
        \hline
    \end{tabular}}
  
    \label{table1}
\end{table}
\s{\textbf{Implementation Details}:
 We utilize ResNet-18 \cite{resnet} as our teacher network, which comprises approximately 11.5 million (11.5m) parameters. As for our student networks, we employ SqueezeNet 1.1, containing around 1.2m parameters. To align the output dimension of SqueezeNet with that of ResNet-18, we append a single-layer fully connected network without an activation function at the end of SqueezeNet. The maximum number of students (i.e $M$) is determined as $\lceil{\frac{\text{N}}{2}}\rceil$; for example, for CIFAR-10, $M$ is set to 3. Both instance projectors and predictors are implemented as 2-layer fully connected neural networks, with dimensions $d\xrightarrow[]{}512\xrightarrow[]{}128$, where $d = 512$ for instance projectors and $d = 128$ for predictors. The cluster projector is designed as a 2-layer fully connected neural network with dimensions $512\xrightarrow[]{}512\xrightarrow[]{}\lambda_t$, where $\lambda_t$ denotes the number of clusters in the current task. It's worth \b{re-emphasizing} that the first layer of the cluster-projector is shared between tasks, and we store the last layer in memory for predicting cluster assignments. In all experiments, the batch size is 256.}

\vspace{-10mm}
\subsection{Comparison Results}\label{comparison}

 To the best of our knowledge, no existing continual learning algorithm is designed explicitly for the clustering task. Therefore, in this section, we compare our proposed FBCC algorithm with state-of-the-art UCL algorithms such as CCL \cite{ccl}, STAM \cite{stam}, LUMP \cite{lump}, CaSSLe \cite{cassle}. Also, we compare our FBCC with two state-of-the art SCL method Co$^2$L \cite{cha2021co2l} and OCD-Net \cite{ocd_net} on three benchmark datasets. Note that Co$^2$L and OCD-Net are supervised methods that make use of data labels during their training phase.  We adopt a common approach followed in the field, as outlined in \cite{SCALE}, by applying the spectral clustering algorithm to the latent representations learned by other algorithms. This approach allows us to evaluate and report $\overline{\text{ACC}}$ and $\overline{\text{F}}$ for these methods. Comparison results are presented in Table \ref{table1}. 

%\vspace{-5mm}
Moreover, we compare our FBCC with a baseline CC \cite{CC} algorithm. CC possesses the flexibility to define its loss function using any pair of samples from distinct clusters, rendering it more potent compared to FBCC, which, in each step, only has access to partial clusters. Consequently, we employ CC as a proxy upper bound for assessing the performance of our algorithm\s{, except for Tiny-ImageNet}.
Note that the lower performance of CC on the Tiny-ImageNet compared to FBCC can be associated with the fact that CC is a memory-hungry algorithm, and running it for more than 256 samples per batch is practically impossible \cite{C3,CC} while the given 256 samples might not be enough for defining relationship between samples when dealing with numerous clusters like the case of Tiny-ImageNet. Yet, CC serves as a proper upper bound for datasets with a small number of classes, such as CIFAR-10 and CIFAR-100.

As illustrated in Table \ref{table1}, FBCC demonstrates notable superiority over alternative algorithms concerning both $\overline{\text{ACC}}$ and $\overline{\text{F}}$. Notably, FBCC surpasses Co$^2$L and OCD-Net, which learns latent representations of data in a supervised manner, and the state-of-the-art UCL algorithm, CaSSLe, while employing fewer parameters. Specifically, FBCC utilizes 15.1m parameters for CIFAR-10 and 17.5m for CIFAR-100 and Tiny-ImageNet, whereas CaSSLe employs 23m parameters across all datasets. The advantage of FBCC stems from two key factors: Firstly, FBCC is tailored for clustering tasks, concurrently learning data sample representations and performing clustering, whereas other UCL algorithms primarily focus on refining data representation. Secondly, FBCC adeptly retains knowledge from multiple previous tasks by leveraging multiple, but a limited number of, specialized student models to mimic the representations acquired by the teacher model for specific tasks.

\vspace{-5mm}
\subsection{Ablation Study}
\label{ablation}
\begin{table}[t]
  \caption{Ablation study of FBCC in terms of $\overline{\text{ACC}} (\%)$ and $\overline{\text{F}} (\%)$. The best result for continual learning algorithms in each column is highlighted in bold.}
    \centering
    \setlength{\tabcolsep}{4pt}
    \renewcommand{\arraystretch}{1.5}
    \scalebox{0.75}{
    \begin{tabular}{|c|c|c|c|c|c|c|c|}
        \hline
        \multirow{2}{*}{Algorithms} &
         \multicolumn{2}{c|}{CIFAR-10} & \multicolumn{2}{c|}{CIFAR-100} & \multicolumn{2}{c|}{Tiny-ImageNet}\\
        \cline{2-7}
       & $\overline{\text{ACC}}$ $(\uparrow)$ & $\overline{\text{F}}$ $(\downarrow)$ &$\overline{\text{ACC}}$ $(\uparrow)$ & $\overline{\text{F}}$ $(\downarrow)$ &$\overline{\text{ACC}}$ $(\uparrow)$ & $\overline{\text{F}}$ $(\downarrow)$\\
        \hline
        % CCL & 36.56 & 6.21 & 19.59& 8.51 & 8.21 & 4.68\\
        % STAM & 39.61  &  5.15 & 25.34  & 6.25 & 9.21 & 4.26 \\
        % LUMP & 56.43 & 12.76 & 19.53& 6.16 & 10.53 & 2.51\\
        % CaSSLe & 40.56 & 3.28 & 36.67 & 3.92 & 17.45 & 2.69\\
        % Co$^2$L (SCL) & 28.35 & 14.05 & 19.88 &  10.30 & 8.69 & 4.95\\
        % \hline \hline
        FBCC w/o Pro & 75.00 & 2.19 & 37.61 & 4.10 & 17.91 & 2.37\\
        FBCC w/o KD & 67.54 & 9.21 &  32.47  & 13.31 & 14.28 &6.58\\
        FBCC + CaSSLe & 70.69  & 4.63 & 35.21  & 6.41& 15.28& 3.01\\
        \hline 
        \textbf{FBCC} & \textbf{75.35} & \textbf{2.12} &   \textbf{38.33} & \textbf{3.81}& \textbf{18.25} & \textbf{2.03}\\
       \hline 
    \end{tabular}}
  
    \label{tb2}
\end{table}
\textbf{Effectiveness of Prototypes in Forward Knowledge Distillation}: To demonstrate the effectiveness of including prototypes learned from the previous task (i.e., $\mathcal{P}_{t-1}$) during training in the forward distillation phase, we propose to exclude prototypes and focus solely on the contrastive loss, i.e., we remove the second term in the denominator of \eqref{lossa_con}.\s{following a formulation similar to that in \cite{SimCLR}.} All other model configurations remain unchanged. This setup is labeled as FBCC w/o Pro in Table \ref{tb2}.

Upon comparing the results obtained from FBCC and FBCC w/o Pro, it is evident that the inclusion of prototypes in forward distillation leads to improved $\overline{\text{ACC}}$ and $\overline{\text{F}}$ across various tasks. This enhancement is attributed to our model's ability to effectively distinguish between data from the current task and prototypes, which serve as representatives of previous tasks.

\textbf{Effectiveness of Students in Forward Knowledge Distillation:} In Table \ref{tb2}, we present a comparison of our proposed method with two alternative configurations in terms of $\overline{\text{ACC}}$ and $\overline{\text{F}}$ . In one of these configurations, denoted as FBCC w/o KD, we exclude the knowledge distillation loss from students to the teacher (i.e., $\mathcal{L}_{dis}$) when updating the parameters of the teacher encoder. In the second configuration, inspired by \cite{cassle}, instead of training multiple students, we employ a strategy where we utilize a previously trained teacher model to mitigate catastrophic forgetting. We freeze the parameters of this copied network, and the knowledge distillation loss is defined in \cite{cassle}. This configuration is labeled as FBCC + CaSSLe in Table \ref{tb2}.

If we compare FBCC w/o KD with FBCC, we observe approximately a 5.88\% improvement in terms of $\overline{\text{ACC}}$ and 7.05\% improvement in terms of $\overline{\text{F}}$ across all datasets. This improvement is primarily attributed to the effectiveness of knowledge distillation from students to the teacher using $\mathcal{L}_{dis}$ in retaining knowledge from previous tasks.

Moreover, upon comparing results obtained from FBCC + CaSSLe with those from FBCC, we can conclude that the effectiveness of having multiple students lies in retaining knowledge from more than one previous task. It is worth noting that the number of parameters for FBCC on CIFAR-10, CIFAR-100, and Tiny-ImageNet are 15.1m, 17.5m, and 17.5m, respectively, while the number of parameters for FBCC + CaSSLe for all datasets is 23m. Our FBCC achieves better results in terms of 3.58\% improvement in $\overline{\text{ACC}}$ and exhibits 2.03\% improvement in terms of $\overline{\text{F}}$ across all datasets despite having fewer parameters.
\vspace{-3mm}
\subsection{Effect of Number of Students in Forward Knowledge Distillation:} \label{changing_m}

\begin{figure}[t]
    \centering
    \includegraphics[width=0.6\textwidth]{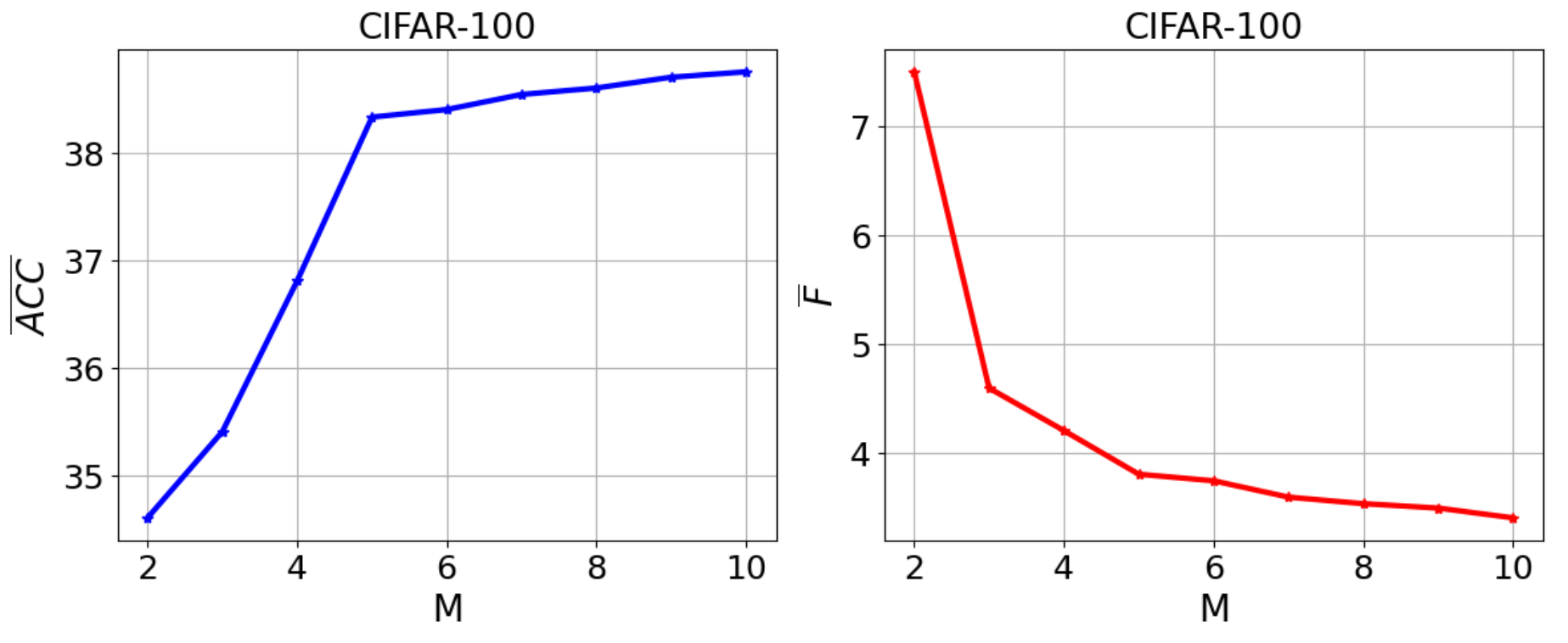} % Replace example-image with your actual image file name
    \caption{Average ACC and Average Forgetting of different values of $M$.}
    \label{fig2}
\end{figure}

In this section, we investigate the effect of the number of students (i.e., $M$) on the performance of FBCC in terms of $\overline{\text{ACC}}$ and $\overline{\text{F}}$. We vary $M$ from 2 to 10 for the CIFAR-100 dataset and report the results in terms of $\overline{\text{ACC}}$ and $\overline{\text{F}}$. The findings are illustrated in Figure \ref{fig2}. As shown in this figure, we observe a significant improvement in performance when changing $M$ from 2 to 5. This improvement is attributed to our model's ability to effectively remember previous tasks. However, when changing $M$ from 6 to 10, we do not observe much improvement. This is because forcing our model to remember numerous previous tasks limits its ability to learn new tasks effectively.

\s{\subsection{Effectiveness of Backward Distillation} \label{BD}

In this section, we delve into the effectiveness of $\mathcal{L}_{stu}$ in transferring knowledge from teacher to student on CIFAR-100 dataset. To achieve this, we define an average over the difference between ACC of the teacher and the ACC of the student. This is defined as follows:
\begin{align}
    \hat{\text{ACC}} = \frac{1}{N} \sum_{t\in \{1,...,N\}} (\text{ACC}_{t,t}^T - \text{ACC}_{t,t}^{S_{l_t}})
\end{align}
where $\text{ACC}_{t,t}^T$ and $\text{ACC}_{t,t}^{S_{l_t}}$ represent the ACC of the teacher and the student on the task $t$ after completing training on the task $t$, respectively. To obtain $\text{ACC}_{t,t}^{S_{l_t}}$ for dataset $\mathcal{D}_t$, after completing training on $\mathcal{D}_t$, we take the output of the student, i.e. $h_{\mathcal{D}_t}^{S_{l_t}}$, and feed it into the cluster projector learned during forward distillation to obtain cluster assignments for the dataset, i.e. $c_{\mathcal{D}_t}= \text{argmax}[C_t(h_{\mathcal{D}_t}^{S_{l_t}})]$. Subsequently, we compare these assignments with the true cluster assignments to compute the $\text{ACC}_{t,t}^{S_{l_t}}$. 

Moreover, we consider three architectures for the student, namely MobileNetV3 Small \cite{Howard2019SearchingFM}, ShuffleNetV2 (0.5x) \cite{ma2018shufflenet}, and SqueezeNet 1.1 \cite{iandola2016squeezenet}. Inspired by \cite{alkhulaifi2021knowledge}, to select the best model for our student, we define a distillation score (DS) that takes into account the size and accuracy of the student relative to those of the teacher network. This score helps us identify the optimal model for the student. The formula for DS is defined as follows:
\begin{align}
    DS = \alpha (\frac{\# Param_S}{\# Param_T}) + (1-\alpha)(1-\frac{1}{N}\sum_{t = 1}^{N} \frac{\text{ACC}_{t,t}^{S_{l_t}}}{\text{ACC}_{t,t}^T}),
\end{align}
where $\# \text{Param}_S$ and $\# \text{Param}_T$ represent the number of parameters of the student and teacher network, respectively, and $\alpha \in [0,1]$ is a hyperparameter that highlights the importance of the first ratio over the second one.  Lower DS values indicate better models. In our experiments, $\alpha$ is set to 0.5 and the teacher network is ResNet-18 with 11.5 million parameters.

\begin{table}[htbp]
  \caption{Comparison of Different Student Architectures in Terms of number of parameters, $\hat{\text{ACC}} (\%)$ and DS. The Best Result in Each Column is Highlighted in Bold}
    \centering
    \setlength{\tabcolsep}{4pt}
    \renewcommand{\arraystretch}{1.5}
    \begin{tabular}{|c|c|c|c|c|}
        \hline
        \multirow{2}{*}{Students} &
         \multicolumn{3}{c|}{CIFAR-100} \\
        \cline{2-4}
       & $\# Param_S$ $(\downarrow)$& $\hat{\text{ACC}}$ $(\downarrow)$ & $\text{DS}$ $(\downarrow)$ \\
        \hline
        MobileNetV3 Small & 2.5m & \textbf{1.05} &  0.120\\
        ShuffleNetV2 (0.5x) & 1.3m & 1.91 & 0.081\\
        SqueezeNet 1.1 & \textbf{1.2m} & 1.68 & \textbf{0.075}\\
               % \hline \hline 
        %FBCC offline &  &  & &  & & \\
        \hline
    \end{tabular}
  
    \label{table2}
\end{table}

Table \ref{table2} shows the comparison of different student architecture in terms of number of parameters, $\hat{\text{ACC}} (\%)$, and DS. By comparing the $\hat{\text{ACC}}$ of students with different architectures, we can infer the effectiveness of $\mathcal{L}_{stu}$ in knowledge distillation from the teacher to the students. For instance, the ACC of SqueezeNet 1.1 with 1.2 million parameters is 1.68\% less than the ACC of ResNet-18 with 11.5 million parameters on average across different tasks on the CIFAR-100 dataset.

Based on DS reported in Table \ref{table2}, we choose SqueezeNet 1.1 as our student network for the CIFAR-100 dataset, as it has the lowest DS among the other architectures. We observe similar pattern for the other datasets as well.}
\vspace{-3mm}
\subsection{Imbalanced Dataset}

In this section, we delve into assessing the efficacy of our proposed FBCC in tackling learning tasks characterized by highly imbalanced sample distributions. To accomplish this, we adopt a strategy wherein we selectively sample data from task $t$ within the CIFAR-10 dataset. Instances from the first task are incorporated into the training set with a likelihood of 0.1, while instances from the final task are included with a likelihood of 1. Instances from intermediate tasks are chosen proportionally, following a linear progression. The inherent challenge posed by imbalanced data lies in the scenario where our model is trained on a limited number of instances from the current task, yet it encounters increasingly more samples from subsequent tasks. This imbalance heightens the risk of CF wherein the model's performance on the current task deteriorates as it learns new tasks, potentially leading to performance degradation in future tasks.

\begin{figure}[t]
    \centering
    \includegraphics[width=0.75\textwidth]{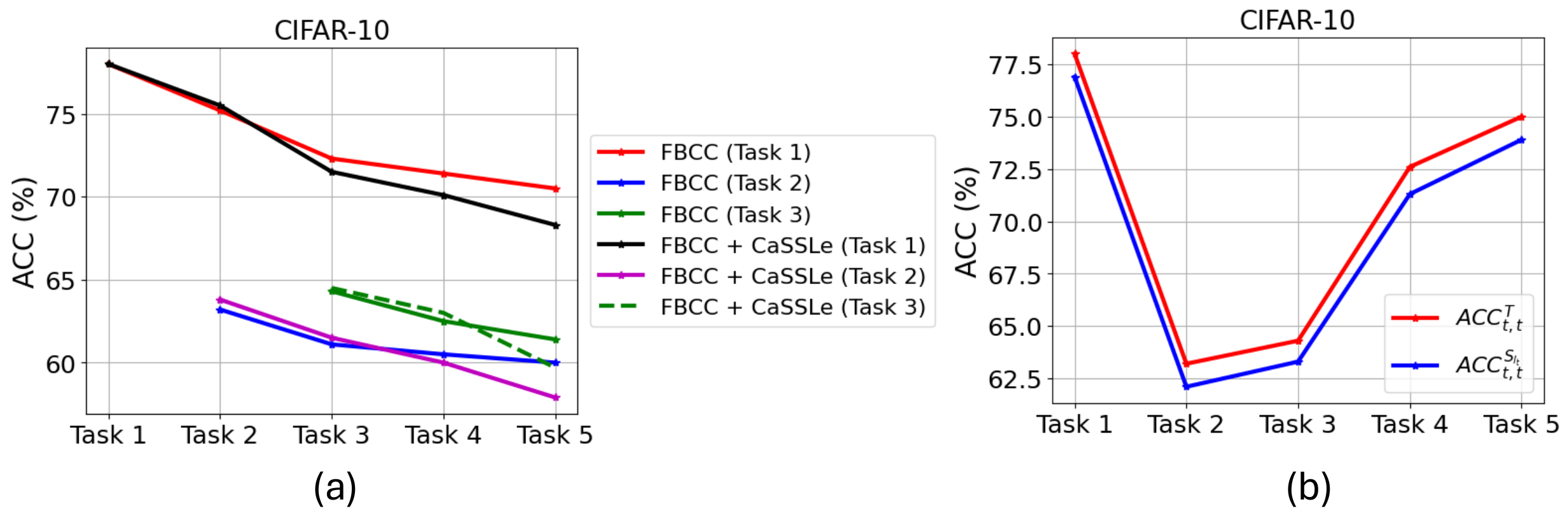} % Replace example-image with your actual image file name
    \caption{Experiments on imbalanced data.}
    \label{fig3}
\end{figure}

In Figure \ref{fig3}-(a), we present a comparative analysis between our proposed FBCC approach and FBCC + CaSSLe. The figure illustrates ACC achieved on the first three tasks ($1\leq t \leq 3$) after completing training on each task. As depicted in the figure, transitioning from task $t$ to task $t+1$ reveals that the performance of FBCC + CaSSLe for task $t$ surpasses that of FBCC. This discrepancy arises due to the utilization of a teacher network explicitly trained on task $t$ within the FBCC + CaSSLe framework. In contrast, FBCC employs a student network with notably fewer parameters for task retention. However, following task $t+2$, a reversal in performance is observed. FBCC exhibits superior performance on task $t$ compared to FBCC + CaSSLe. This shift can be attributed to FBCC's utilization of multiple specialized student networks, thereby enhancing its capacity to retain knowledge from previous tasks effectively. For example, this characteristic becomes particularly evident when examining the performance on task 1. At the conclusion of training on task 2, FBCC + CaSSLe exhibits superior performance compared to FBCC. However, after completing task 3, the trend reverses, with FBCC surpassing FBCC + CaSSLe in terms of ACC. 

Furthermore, to demonstrate the effectiveness of training students to mimic the behavior of the teacher network on imbalanced data, we plot $\text{ACC}_{t,t}^{\text{T}}$ and $\text{ACC}_{t,t}^{\text{S}_{l_t}}$, as discussed in Section 3 of the supplementary file, for imbalanced data in Figure \ref{fig3}-(b). For our experiments, we utilize SqueezeNet 1.1 \cite{squeezenet}. As depicted in the figure, our student network adeptly follows the teacher network in generating quality representations for the each task on imbalanced datasets.

\vspace{-5mm}

\section{Conclusion}
\vspace{-3mm}
In conclusion, while UCL shows promise for sequential learning without explicit label information, the lack of unsupervised continual clustering methods poses a significant challenge, especially with the phenomenon of CF where models forget previous tasks. Existing strategies for handeling CF like knowledge distillation and replay buffers have limitations. To address this, we propose Forward-Backward Knowledge Distillation for Continual Clustering (FBCC). FBCC employs a singular continual learner (``teacher'') with a cluster projector and multiple student models to combat CF effectively. Through Forward and Backward Knowledge Distillation phases, FBCC enables the teacher to learn new clusters while retaining past knowledge, empowering student models to mimic the teacher's behavior. Our experiments demonstrate FBCC's efficacy compared to existing methods, offering a promising solution for real-world applications in dynamic environments. FBCC presents a novel approach to continual clustering within UCL, preserving memory efficiency and paving the way for advancements in machine learning applications in evolving contexts.
\s{In conclusion, UCL presents a promising avenue for enabling neural networks to learn sequentially without explicit label information. However, the absence of continual clustering methods within UCL poses a significant challenge, exacerbated by the phenomenon of CF, where models forget previously learned tasks when learning new ones. While existing strategies like knowledge distillation and replay buffers aim to mitigate CF, they often encounter issues of memory inefficiency or privacy concerns. In response to this challenge, we propose Forward-Backward Knowledge Distillation for Continual Clustering tasks (FBCC). FBCC introduces a singular continual learner (``teacher'') with a cluster projector, supported by multiple student models, to effectively address CF. By employing two distinct phases – Forward Knowledge Distillation and Backward Knowledge Distillation – FBCC enables the teacher to learn new clusters while retaining knowledge from past tasks and empowers student models to mimic the teacher's behavior, ensuring retention of task-specific knowledge for future training iterations. Our experimental results underscore the efficacy of FBCC compared to existing methods, underscoring its potential for practical deployment in dynamic environments. Through FBCC, we contribute a novel approach to continual clustering within UCL, offering a promising solution to the challenge of CF while preserving memory efficiency, thereby paving the way for advancements in real-world applications of machine learning in dynamic and evolving contexts.}

\bibliographystyle{splncs04}
\bibliography{main}
\end{document}